\documentclass{llncs}
\usepackage{graphicx}
\usepackage{amsmath}
\usepackage{adjustbox}
\usepackage[utf8]{inputenc}
\usepackage{csquotes}
\begin{document}

\title{Coin\_flipper at eHealth-KD Challenge 2019}

\subtitle{Voting LSTMs for Key Phrases and Semantic Relation Identification Applied to Spanish eHealth Texts}
\titlerunning{Voting LSTMs}        

\author{Neus Català\inst{1}\orcidID{0000-0002-6184-0367} \and
Mario Martin\inst{2}\orcidID{0000-0002-4125-6630}}

%

\institute{TALP Research Center - Universitat Politècnica de Catalunya, Spain\\
\email{ncatala@talp.upc.edu} \and
Barcelona Supercomputing Center and
Computer Science Department at Universitat Politècnica de Catalunya, Spain\\
\email{mmartin@cs.upc.edu}}
\maketitle              
{\let\thefootnote\relax\footnotetext{Copyright \textcopyright\ 2019 for this paper by its authors. Use permitted under Creative Commons License Attribution 4.0 International (CC BY 4.0). IberLEF 2019, 24 September 2019, Bilbao, Spain. }}

\begin{abstract}
This paper describes our approach presented for the eHealth-KD 2019 challenge.
Our participation was aimed at testing how far we could go using generic tools for Text-Processing but, at the same time, using common optimization techniques in the field of Data Mining. The architecture proposed for both tasks of the challenge is a standard stacked 2-layer bi-LSTM. The main particularities of our approach are: (a) The use of a surrogate function of \textit{F1} as loss function to close the gap between the minimization function and the evaluation metric, and (b) The generation of an ensemble of models for generating predictions by majority vote. Our system ranked second with an \textit{F1} score of 62.18\% in the main task by a narrow margin with the winner that scored 63.94\%.

\keywords{Key phrase detection \and detection of semantic relations \and LSTMs \and Majority Voting \and F1 loss function}
\end{abstract}

\section{Introduction}

This article describes the model presented by the \textit{coin-flipper} team for solving the shared task presented in eHealth-KD 2019 challenge \cite{ehealthkd19_overview}. The main goal of the challenge is to encourage the development of software  systems to automatically extract information from electronic health documents written in  Spanish. The challenge involves two subtasks: 1) Identification and classification of key phrases (four different types of units) and 2) Detection of semantic relations among them (thirteen semantic relations).

The tasks proposed are similar to previous competitions such as Semeval-2017 Task 10: ScienceIE \cite{augenstein-etal-2017-semeval} and TASS-2018-Task 3 eHealth Knowledge
Discovery \cite{DBLP:conf/sepln/2018tass}, but the types of entities and the amount of semantic relations to be identified have been changing over time. Also, we find changes in the systems approaches evolving from rule based methods to Deep Learning models some of them incorporating domain-specific knowledge.

As the organizers of the challenge state,
although this challenge is oriented to the health domain, the kind of knowledge to be extracted is general-purpose. With this idea in mind, we propose a model that avoids the use of knowledge or tools specific to the domain, easing the portability of the model to new domains and tasks.

\section{Tasks and data}

The eHealth-KD 2019 challenge involves two subtasks: subtask A) Identification and classification of key phrases (four different types of units), and subtask B) Detection of semantic relations among them (thirteen semantic relations).

Each subtask has been evaluated individually and also as components of a pipeline. For this purpose, three evaluation scenarios have been proposed. Scenario 1 requires subtasks A and B to be executed sequentially in a pipeline. Scenario 2 only evaluates subtask A and Scenario 3 only evaluates subtask B.

A training set that contains a total of 600 sentences was provided to be used for the learning step. For the validation step, an additional set of 100 sentences was available.

A detailed explanation of the corpus data, subtasks and evaluation metrics can be found in the overview description paper \cite{ehealthkd19_overview}.

\section{Our approach}

Our aim was to see how far we can get in tasks described above using generic NLP tools (non-specific for the domain) and standard Data Mining strategies.

About the model to be learned, given the sequential nature of the data (sentences), we proposed a stacked 2-layer bi-LSTM architecture. LSTMs have been proved a successful approach in several Natural Language Processing tasks, such as Named Entity Recognition (f.i.~\cite{huang2015bidirectional}) and Semantic Role Labeling (f.i.~\cite{zhou-xu-2015-end}). About the data mining strategies, we applied two following strategies:

\begin{enumerate}
    \item Building an ensemble of models for generating predictions:

    It is well known in Machine Learning that ensemble models have higher performance than single models~\cite{Zhou:2012:EMF:2381019}, specially when models have a high variance which is the case of neural networks. So, for each task, we build an ensemble of models instead of a single model.

    There are several ways to generate an ensemble of neural networks. Usually, differences in hyperparameters or different architectures are enough to ensure diversity of models necessary for making ensemble methods profitable. In our case we achieve diversity by using random initialization of models before training~\cite{Goodfellow2015}.

    There also exist a lot of different techniques to combine the output of different models into a single prediction~\cite{Zhou:2012:EMF:2381019}. We will apply the simplest method of
    majority vote to combine the outputs of the models in a ensemble.

\item The use of a surrogate function of \textit{F1} metric as the loss function:

Usually, accuracy is the metric to be optimized and Cross-entropy (CE) is the continuous function used as loss function. However, in this challenge, \textit{F1} score is used for evaluation. The use of CE is not optimal in this case because we would like to minimize the error according to the metric used for evaluation. In~\cite{Eban2017}, authors propose to solve this problem by defining continuous surrogate upper and lower bounds for true positive and false positive cases and, from them, \textit{F1'} surrogate. Then they use $1/{F1'}$ as the loss function to minimize. We have done it differently: in our case, getting the predictions $\hat{y}$ from our Neural Network (with \textit{sigmoid} function activation in last layer), we choose to use a soft version of \textit{F1} metric. We define soft-true positive cases ($tp_s$) and soft-false positives ($fp_s$), where labels $y$ are in \{0,1\} and predictions $\hat{y}$ in [0..1], as:

\begin{align*}
tp_s & =  \sum_{i \in Y^+} \hat{y}_i =  \sum_{i \in Y} y_i \hat{y}_i \\
fp_s & =  \sum_{i \in Y^-} \hat{y}_i =  \sum_{i \in Y} (1-y_i)\hat{y}_i
\end{align*}

\noindent
Notice that when predicted labels $\hat{y}$ are hard (0 or 1), these values are exactly true and false positive cases, respectively.

Replacing soft versions of true and false positives cases in \textit{F1} definition (in the type I and type II error form), we have our surrogate $F1_s$ function to maximize.
\begin{equation}
    F1_s \; = \; \frac{2 \; tp_s}{\left|Y^+\right|+tp_s+fp_s} \; = \;
\frac{2\;  \sum_{i\in Y^+} \hat{y}_i}{\left|Y^+\right| + \sum_{i\in Y^+} \hat{y}_i+ \sum_{i\in Y^-} \hat{y}_i}
\end{equation}

\noindent
Note that maximize $F1_s$ implies increasing $\hat{y}$ of positive cases while decreasing $\hat{y}$ of negative cases in a proportional way to \textit{F1} function. Loss function to minimize will be $1-F1_s$.

\end{enumerate}{}

In the next sections we describe further details about how subtasks A and B are solved.

\subsection{Subtask A}

Subtask A consists in detecting key phrases in texts (from now on we will abbreviate key phrase to KPhr) and their kind or class. For this task we set as input only the sequence of words of each sentence and their \textit{Part-of-Speech} (PoS).
Words are represented using \textit{FastText} embeddings \cite{bojanowski-etal-2017-enriching} trained on the \textit{Spanish Billion Word Corpus} \cite{cardellinoSBWCE}. Among the advantages of \textit{FastText} over other approaches are that, being \textit{FastText} trained on n-grams of words, embeddings carry implicitly morphological information and also are able to output a reasonable embedding for a word not seen during training.

PoS for sentences are generated using \textit{spaCy}\cite{spacy2} tagger for Spanish. Probably better results would be obtained using more advanced taggers like \textit{Free\-Ling} PoS tagger \cite{padro12}.

In the definition of the output of the architecture we were inspired on image segmentation work in which, given an image, a mask over the image is generated that encodes for each pixel the class it belongs to. Similarly, for each sentence, we define as output a mask vector where for each word we have codified the kind of KPhr the word belongs to (or a label denoting that the word does not belong to any KPhr).

The architecture of the model to be learned consists of two stacked layers of bi-LSTM feed with the word embedding and the one-hot encoding of the PoS for each word of the sentence. We apply sequence Padding (with length parameter equal to the length of the larger sentence in data set) and Masking. Initially, output was defined as a vector with the one-hot codification of the corresponding KPhr label for each word.
However, we found empirically that learning a model to predict \textit{all} kinds of KPhr was not very successful. So, we decided to split the task into learning models specialized in predictions for each kind of KPhr. With this modification, now the output of the architecture is a 0/1 vector with 1 denoting if a word belongs to the kind of KPhr the model is trained for, and 0 otherwise. Notice that with this modification, the use of \textit{F1} loss become even more important because the induced sparsity of labels.

\begin{figure*}[t!]
   \centering
    \includegraphics[trim={0 0cm 13cm 1cm},width=9.5cm,clip]{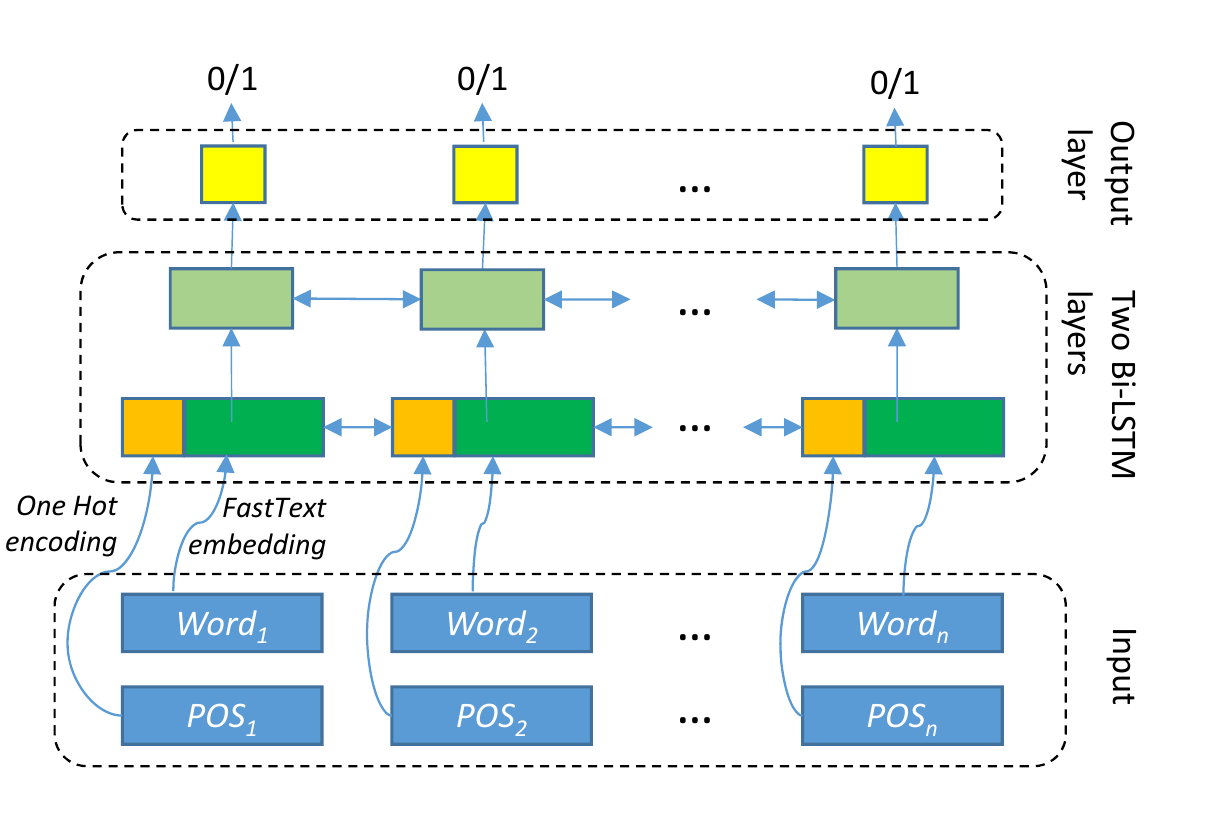}
   \caption{Basic two level bi-LSTM architecture for subtask A. Input are words and their PoS that are embedded using \textit{FastText} and one-hot encoding, respectively. For each kind of key phrase we train a separate ensemble of 15 models. Output in subtask A is 0/1 with 1 denoting the current word belongs to a key phrase of the kind we are training for, and 0 otherwise.}
   \label{figura1}
\end{figure*}

The learning of each model is done by using surrogate $1-F1_s$ loss (as explained in previous section) with early stopping when $F1$ score does not improve in 100 epochs on the development dataset, and returning the model that reaches the best $F1$ score on development dataset. Default learning rate $0.001$ with \textit{adam} optimization is used. Batch size is 32 sentences. Neither dropout nor data augmentation is used. First bi-LSTM layer has 150 units in each direction. Second bi-LSTM layer has 32 units in each direction. The same architecture is used for all KPhr detection models without fine-tuning of parameters. See a graphical representation of the model in Figure~\ref{figura1}.

As stated in the previous section, we create an ensemble of 15 models for each target KPhr that is used later to generate the final prediction by majority voting. Diversity of models in the ensemble is obtained only from different initialization of weights. To avoid the vote of specially bad models, each model is applied to the development dataset and those models with an \textit{F1} score far below the average of the ensemble are removed from it.

Our approach has two obvious problems. The first one is that, because the way we define segmentation, we don't know when a KPhr starts nor ends. A naive solution would be to join into the same KPhr consecutive words that are tagged with the same label. However this heuristic solution does not always work. We solve this problem by inspecting the PoS of consecutive words tagged with the same label. For each combination of PoS found, we compute how many cases covered are in fact cases where words belong to the same KPhr. From the statistics of these cases we create a set of rules that decide if two consecutive words with the same label really belong to the same KPhr or not.

The second problem is that having different models, one for each kind of KPhr, words could be labeled as belonging to different kinds of KPhr at the same time. So, in case of conflict among different ensembles, in order to decide the final kind of KPhr, we count how many positive votes each ensemble has emitted and multiply it by a weighting parameter. The ensemble with the highest product is the winner and the word is labeled with the kind of KPhr the ensemble represents. Weights for each ensemble are found by applying the ensembles on the development dataset and grid-searching for the combination of weights that returns highest global \textit{F1}.

\begin{figure*}[t!]
   \centering
    \includegraphics[trim={13cm 0 0cm 0},width=9.5cm,clip]{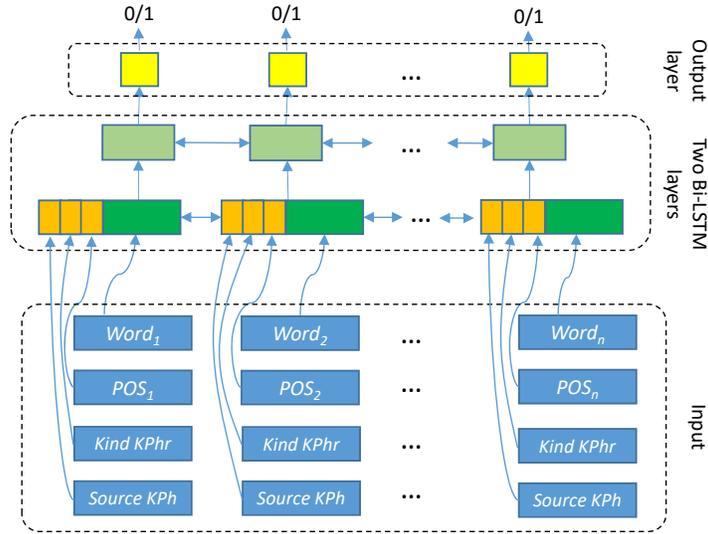}
\medskip
   \caption{Two layer bi-LSTM architecture for subtask B detecting relation \textit{rel}. Given a \textit{source KPhr} (identified in subtask A), output consists of 0/1 with 1 denoting if the current word belongs to a KPhr that is \textit{target} of \textit{rel} for the \textit{source KPhr}. Input consists of (for each word of the sentence) the word, PoS tag, kind of KPhr the word belongs to (if any) and, finally, if current word belongs to \textit{source KPhr} or not. Words and PoS are embedded as in subtask A, and all other information is embedded using one-hot encoding.}
   \label{figura2}
\end{figure*}

\subsection{Subtask B}

Subtask B consists in detecting semantic relations (thirteen types of relations) between KPhr pairs identified in the previous task.
For this task, given a relation \textit{rel} and a KPhr considered as \textit{source} for the relation, we try to identify another KPhr in the sentence with the role \textit{target} for this relation \textit{rel}. In order to learn models able to solve this task, we
define an architecture that, given a relation \textit{rel} and a \textit{source KPhr}, has the following inputs:

\begin{enumerate}
    \item Sequence of words in the sentence where \textit{source KPhr} is found.
    \item Part of Speech tags for each word in the sentence.
    \item Mask denoting for each word the \textit{kind of KPhr it belongs to} (codifying in 1-4 the kind of KPhr and 0 if it does not belong to any KPhr).
    \item 0/1 Mask vector for each word with 1 denoting if the word belongs to the \textit{source KPhr}, and 0 otherwise.
\end{enumerate}

Words are represented as \textit{FastText} vector embeddings and all other inputs are encoded using one-hot encoding. All vectors codifying the information of one word are concatenated. See Figure~\ref{figura2} for a representation of the architecture.

For each relation we build a Neural Network that consists of the embedding layers explained above plus 2 stacked bi-LSTM layers. As in subtask A, we use padding and masking. The output
consists of a mask vector 0/1 with 1 only for those words who belong to a \textit{target} KPhr of the relation \textit{rel} of current KPhr as source. When no \textit{target} KPhr exists for the current relation, output is all zeros.

When the model identifies at least one word of a target KPhr, we consider the whole KPhr detected. The output representation selected allows that in the same sentence one KPhr as \textit{source} has several KPhr as \textit{target} for the same relation.

All training parameters and procedures are the same as those used in subtask A.
Again, the learning of each model is done by using surrogate $1-F1_s$ loss with early stopping when $F1$ score does not improve in 100 epochs on the development dataset, and returning the model that reaches the best $F1$ score on development dataset. Default learning rate $0.001$ with \textit{adam} optimization is used. Batch size is 32. Neither dropout nor data augmentation is used. First bi-LSTM layer has 150 units in each direction. Second bi-LSTM layer has 32 units in each direction. The same architecture is used for all relations without fine-tuning of parameters. For each relation, an ensemble of 15 models is built from which majority voting is implemented.

In some cases, ensembles of different relations identify the same pair of KPhr as positive cases. When this happens, we assign to the pair of KPhr the relation of the ensemble with highest \textit{precision} on the development dataset.

\section{Results}

Table \ref{table2} shows \textit{F1} scores for the three scenarios achieved by the baseline system provided by organizers, the best method and our method.

Our results were very competitive. We were second in the main evaluation (scenario 1) by a narrow margin - only 1.76 of \textit{F1} score over 100. In scenario 2 and, specially, scenario 3 we were at a larger distance to the best method (see Table~\ref{table2}) in positions 5th and 4th of the ranking, respectively. This is surprising because scenario 1 is the composition of the other 2 scenarios. The main difference in evaluation among scenarios was that for scenario 1 a subset of 100 sentences out of a 8.800 sentences dataset was randomly selected for evaluation, while in scenarios 2 and 3 a dataset of 100 sentences was given to each one for evaluation.

For scores shown in Table~\ref{table2} we did not submit results for different para\-me\-ters except for subtask B. In subtask B, we noticed that not all relations had the same difficulty in being detected. We found that a first set of 6 relations (\textit{in-context}, \textit{subject}, \textit{target}, \textit{domain}, \textit{arg} and \textit{is-a}) were successfully identified with high recall and precision values when tested on development dataset (above 75\%). Relations \textit{in-place}, \textit{has-property}, \textit{same-as} and \textit{in-time} were (in this order) harder to identify and, finally, for \textit{part-of}, \textit{causes} and \textit{entails} relations, results were so bad that models for those relations were discarded.

In order to obtain the highest score, we ranked relations by their \textit{F1} scores on development dataset and submitted different sets of results considering only relations that were in 6, 7, 8, 9 and 10 top positions. Best results on testing dataset were obtained considering top 7 relations, that is, the ones in the first set (\textit{in-context}, \textit{subject}, \textit{target}, \textit{domain}, \textit{arg} and \textit{is-a}) plus \textit{in-place}. Considering only a subset of possible relations had an obvious impact on final performance that achieved good precision but low recall.

We think that difficulty in detecting relations \textit{part-of}, \textit{same-as}, \textit{in-time}, \textit{entails} and \textit{has-property} comes from the low number of instances (range 50-100) in the training set. So, for these cases, some techniques like data augmentation should be used to obtain better results. An exception is relation \textit{causes} that had 230 instances and was very difficult to identify with our approach. In this case most complex architectures should be used.

\begin{table} [t!]
  \caption{\label{table2}F1 scores, Precision and Recall obtained for the three scenarios achieved by baseline system, best method presented and our method.}
\begin{adjustbox}{width=\columnwidth}
\begin{tabular} {l | c c c | c c c | c c c}
\hline\rule{-2pt}{8pt}
& \multicolumn{3}{c|}{baseline} &\multicolumn{3}{c|}{\footnotesize{ Joint-BERT-RCNN} (best)}& \multicolumn{3}{c}{Voting LSTMs (ours)}\\
& F1 & P & R & F1 & $\;\;$P & R & F1 & P & R\\
\hline\rule{-2pt}{10pt}
\textit{scenario 1}&43.09&52.04&36.77& 63.94 & $\;\;$65.06 & 62.86 & 62.18 & 74.54 &	53.34\\
\textit{scenario 2}&54.66&51.29&58.51& 82.03 & $\;\;$80.73 & 83.36 & 78.73 & 79.86 & 	77.63\\
\textit{scenario 3}&12.31&48.78&7.04& 62.69 & $\;\;$66.67 &  59.15 & 49.31 & 71.33 & 	37.68\\
\hline
 \end{tabular}
\end{adjustbox}
 \end{table}

\section{Conclusions}
In this paper we have presented a simple approach to solve the tasks proposed in this competition. The aim was to measure how far we can get with standard tools and simple algorithms but using common optimization techniques in the Data Mining field. In particular, the architecture proposed for both tasks was a two level bi-LSTM to modelize each of the classes we wanted to capture. We didn't use state of the art language models like BERT, ELMO or GPT, neither specialized NER techniques, knowledge bases or taggers specific for the domain (so results probably will be similar in other domains).

The main particularities of our approach are: (a) The use of a surrogate loss function of \textit{F1} metric to close the gap between the minimization function and the evaluation metric and (b) the generation of an ensemble of models for generating predictions by majority vote.

From results, we conclude that it is worth using the evaluation metric (or a surrogate function of it) as loss function and also building ensemble models for generating predictions. We did not do an exhaustive study of the improvement in performance due to the use of a surrogate function of \textit{F1} as loss, but in
initial experiments we found a clear advantage of the surrogate function usage instead of standard loss functions.
We neither did an exhaustive study of the impact in performance of learning ensemble models instead of learning single models, however, we noticed that usually \textit{F1} scores obtained by the majority vote of the ensemble was about 1-2 points higher than the best model of the ensemble and 3-4 points higher than the average of the models of the ensemble. Note that combination of outputs in the ensemble is made through simple majority voting but more complex ensemble techniques could improve the performance of the system even more. For a review of ensemble methods see~\cite{Zhou:2012:EMF:2381019}.

The results show that our model achieves good precision scores for the three scenarios but sometimes at the expense of lower recall levels than other systems, specially notorious in scenario 3. The difficulties our model had in identifying some semantic relations may be due to the low number of examples of these relations in the training dataset. To deal with this problem, we are planning to use data augmentation techniques in order to increase the number of instances of the less frequent relations. One of the techniques we want to explore is backtranslation, used in text classification \cite{Shleifer2019}.

\section*{Acknowledgments}
\textbf{Funding}: Neus Català's work has been partially funded by the Spanish Government and by the
European Union through GRAPH-MED project (TIN2016-77820-C3-3-R) and AEI/FEDER, UE. 

Mario Martin's work has been partially supported by the Joint Study Agreement no. W156463 under the IBM/BSC Deep Learning Center agreement.

\bibliographystyle{splncs04}
\bibliography{article}

\end{document}